\documentclass{article}

\usepackage{microtype}
\usepackage{graphicx}
\usepackage{subcaption}
\usepackage{booktabs}   % toprule, midrule, bottomrule
\usepackage{amsmath}    % equation 환경, mathrm% for professional tables

\usepackage{hyperref}

\usepackage[preprint]{icml2026}

\usepackage{amsmath}
\usepackage{amssymb}
\usepackage{mathtools}
\usepackage{amsthm}

\usepackage[capitalize,noabbrev]{cleveref}

\theoremstyle{plain}

\theoremstyle{definition}

\theoremstyle{remark}

\usepackage[textsize=tiny]{todonotes}

\icmltitlerunning{TriFit: Trimodal Fusion with Protein Dynamics for Mutation Fitness Prediction}

\begin{document}

\twocolumn[
  \icmltitle{TriFit: Trimodal Fusion with Protein Dynamics for Mutation Fitness Prediction}



  \icmlsetsymbol{equal}{*}

  \begin{icmlauthorlist}
    \icmlauthor{Seungik Cho}{yyy}
  \end{icmlauthorlist}

  \icmlaffiliation{yyy}{Department of Physics and Astronomy, Rice University, Texas, USA}

  \icmlcorrespondingauthor{Seungik Cho}{seungikcho@rice.edu}
  \icmlkeywords{Machine Learning, ICML}

  \vskip 0.3in
]



\printAffiliationsAndNotice{}  % no special notice (required even if empty)

\begin{abstract}
Predicting the functional impact of single amino acid substitutions (SAVs) is central to understanding genetic disease and engineering therapeutic proteins. While protein language models and structure-based methods have achieved strong performance on this task, they systematically neglect protein dynamics—residue flexibility, correlated motions, and allosteric coupling are well-established determinants of mutational tolerance in structural biology, yet have not been incorporated into supervised variant effect predictors. We present \textbf{TriFit}, a multimodal framework that integrates sequence, structure, and protein dynamics through a four-expert Mixture-of-Experts (MoE) fusion module with trimodal cross-modal contrastive learning. Sequence embeddings are extracted via masked marginal scoring with ESM-2 (650M); structural embeddings from AlphaFold2-predicted $C_\alpha$ geometries; and dynamics embeddings from Gaussian Network Model (GNM) B-factors, mode shapes, and residue-residue cross-correlations. The MoE router adaptively weights modality combinations conditioned on the input, enabling protein-specific fusion without fixed modality assumptions. On the ProteinGym substitution benchmark (217 DMS assays, 696k SAVs), TriFit achieves AUROC $0.897 \pm 0.0002$, outperforming all supervised baselines including Kermut ($0.864$) and ProteinNPT ($0.844$), and the best zero-shot model ESM3 ($0.769$). Ablation studies confirm that dynamics provides the largest marginal contribution over pairwise modality combinations, and TriFit achieves well-calibrated probabilistic outputs (ECE $= 0.044$) without post-hoc correction.
\end{abstract}

\section{Introduction}
Predicting whether a single amino acid substitution (SAV) disrupts protein function is a fundamental challenge in computational biology, with applications ranging from genetic disease diagnosis to therapeutic protein engineering~\cite{fowler2014deep, notin2023proteingym}. Despite a surge in machine learning-based approaches—including sequence-based protein language models~\cite{lin2023esm2, meier2021esm1v}, structure-based inverse folding models~\cite{hsu2022esmif, dauparas2022proteinmpnn}, and hybrid methods combining both~\cite{notin2022trancepteve, su2023saprot, jumper2021alphafold}—a critical third source of biophysical information remains overlooked: \textit{protein dynamics}. Residue flexibility, correlated motions, and allosteric coupling—captured by elastic network models such as the GNM~\cite{bahar1997gnm, haliloglu1997anm}—are well-established determinants of mutational tolerance, yet have not been incorporated into any supervised variant effect predictor. Furthermore, existing fusion strategies treat all modalities uniformly, whereas the relative informativeness of sequence, structure, and dynamics varies considerably across mutation types, motivating an adaptive routing mechanism.

In this work, we present \textbf{TriFit} (\textbf{Tri}modal \textbf{Fit}ness predictor), integrating all three modalities through a Mixture-of-Experts (MoE) fusion module with trimodal contrastive learning. Our key contributions are:
\begin{itemize}
    \item We introduce \textbf{protein dynamics as a third modality} for variant effect prediction, extracting GNM-based B-factors, mode shapes, and residue-residue cross-correlations as per-residue embeddings from AlphaFold2 structures.

    \item We propose an \textbf{input-conditioned MoE fusion} module with four specialized experts (Seq+Struct, Seq+Dyn, Struct+Dyn, Trimodal) and a learned router that adaptively weights modality combinations based on the projected multimodal representation.

    \item We apply \textbf{trimodal cross-modal contrastive learning} across all three modality pairs simultaneously via InfoNCE loss~\cite{oord2018infonce}, aligning complementary representations in a unified latent space.

    \item On the \textbf{ProteinGym substitution benchmark}~\cite{notin2023proteingym} (217 DMS assays, 696k SAVs), TriFit achieves AUROC $0.897 \pm 0.0002$, surpassing all supervised baselines including Kermut (AUROC $0.864$) and ProteinNPT~\cite{notin2023proteinnpt} (AUROC $0.844$), and the best zero-shot model ESM3~\cite{hayes2024esm3} (AUROC $0.769$).
\end{itemize}

\section{Method}

\begin{figure*}[t]
    \centering
    \includegraphics[width=\textwidth]{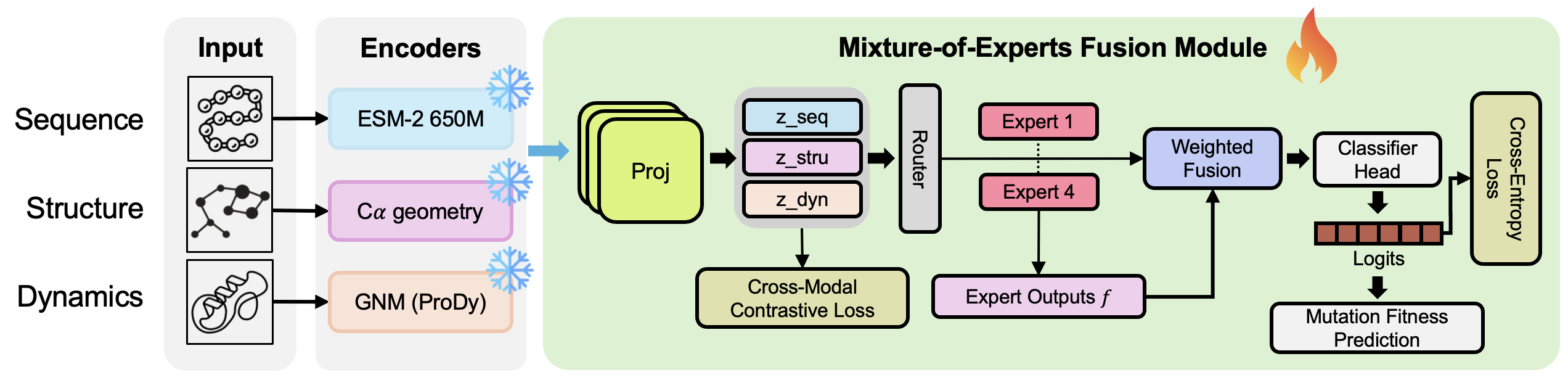}
    \caption{TriFit architecture. Sequence, structure, and dynamics encoders (frozen) extract modality-specific embeddings. A learned projection maps each to a shared 512-dim space. The four-expert MoE router adaptively combines modality pairs (E1: Seq+Struct, E2: Seq+Dyn, E3: Struct+Dyn, E4: Trimodal) via soft gating. Cross-modal contrastive loss aligns all three modality pairs during training. The weighted-fused representation $\mathbf{f}$ is passed to a binary classifier predicting functional vs.\ damaging variants.}
    \label{fig:arch}
\end{figure*}

Figure~\ref{fig:arch} illustrates the overall TriFit architecture. Three modality-specific encoders (frozen) extract per-residue embeddings, which are projected to a shared space and fused through a four-expert MoE module. Cross-modal contrastive loss aligns the three modality representations during training, and the fused representation is passed to a binary classifier for fitness prediction.

\subsection{Problem Formulation}

Given a wild-type protein sequence $\mathbf{s} \in \mathcal{A}^L$ of length $L$ over the amino acid alphabet $\mathcal{A}$, and a single amino acid variant (SAV) $v = (i, a_i \rightarrow a_i')$ that substitutes residue $a_i$ at position $i$ with $a_i'$, we aim to predict a binary fitness label $y \in \{0, 1\}$, where $y=1$ denotes a functional (fit) variant and $y=0$ denotes a damaging variant. Labels are derived from experimentally measured DMS fitness scores via top/bottom 30\% thresholding~\cite{notin2023proteingym}.

\subsection{Multimodal Embedding Extraction}

TriFit extracts per-variant embeddings from three complementary modalities. All modality encoders are kept \textit{frozen}; only the downstream fusion module and classifier are trained.

\paragraph{Sequence Embedding.}
We adopt the masked marginal scoring strategy of ESM-1v~\cite{meier2021esm1v}, which produces contextualized representations without mutant-specific forward passes. For each variant $(i, a_i \rightarrow a_i')$, we mask position $i$ in the wild-type sequence and perform a single forward pass through ESM-2 (650M)~\cite{lin2023esm2} to obtain the context embedding $\mathbf{h}_i^{\text{seq}} \in \mathbb{R}^{1280}$. The final sequence embedding is:
\begin{equation}
    \mathbf{e}^{\text{seq}} = \mathbf{h}_i^{\text{seq}} + \left(\mathbf{t}_{a_i'} - \mathbf{t}_{a_i}\right),
\end{equation}
where $\mathbf{t}_a \in \mathbb{R}^{1280}$ denotes the token embedding of amino acid $a$ extracted from the ESM-2 embedding table. This formulation captures both the structural context of position $i$ and the amino acid substitution signal, requiring only $L_{\text{protein}}$ forward passes per protein regardless of the number of variants.

\paragraph{Structure Embedding.}
We use AlphaFold2-predicted structures~\cite{jumper2021alphafold} provided by the ProteinGym benchmark. For each protein, we construct a $k$-nearest neighbor graph over $C_\alpha$ atoms and extract per-residue geometric features comprising inter-residue distances and direction vectors to the $k=20$ nearest neighbors. These features are projected to $\mathbb{R}^{512}$ via a fixed random projection matrix $\mathbf{W} \in \mathbb{R}^{83 \times 512}$, yielding the structural embedding $\mathbf{e}^{\text{str}} \in \mathbb{R}^{512}$ at the mutation site.

\paragraph{Dynamics Embedding.}
We compute protein dynamics features using the Gaussian Network Model (GNM)~\cite{bahar1997gnm} applied to each AlphaFold2 structure via ProDy~\cite{bakan2011prody}. For each protein, we calculate: (i) normalized B-factors $\mathbf{b} \in \mathbb{R}^L$ (residue flexibility); (ii) the top-$K$ GNM mode shapes $\mathbf{U} \in \mathbb{R}^{L \times K}$ ($K{=}20$); and (iii) mode-projected cross-correlations $\mathbf{C} \in \mathbb{R}^{L \times K}$. Additionally, diagonal stiffness values from the Kirchhoff matrix provide a complementary rigidity measure. These four feature types are concatenated per residue and projected to $\mathbf{e}^{\text{dyn}} \in \mathbb{R}^{256}$ via a fixed random projection.

\subsection{Mixture-of-Experts Fusion Module}

Let $\mathbf{z}^m = \text{Proj}_m(\mathbf{e}^m)$ denote the projected embedding of modality $m \in \{\text{seq, str, dyn}\}$, where each $\text{Proj}_m$ is a learnable MLP followed by LayerNorm, mapping to a common dimension $d{=}512$.

We define four specialized experts that process distinct modality combinations:
\begin{align}
    \mathbf{f}_1 &= \text{Expert}_1([\mathbf{z}^{\text{seq}}; \mathbf{z}^{\text{str}}]), \quad
    \mathbf{f}_2 = \text{Expert}_2([\mathbf{z}^{\text{seq}}; \mathbf{z}^{\text{dyn}}]), \notag \\
    \mathbf{f}_3 &= \text{Expert}_3([\mathbf{z}^{\text{str}}; \mathbf{z}^{\text{dyn}}]), \quad
    \mathbf{f}_4 = \text{Expert}_4([\mathbf{z}^{\text{seq}}; \mathbf{z}^{\text{str}}; \mathbf{z}^{\text{dyn}}]),
\end{align}
where $[\cdot;\cdot]$ denotes concatenation and each expert is a two-layer MLP with GELU activation. The router produces soft assignment weights:
\begin{equation}
    \mathbf{w} = \text{Softmax}\left(\text{Router}\left([\mathbf{z}^{\text{seq}}; \mathbf{z}^{\text{str}}; \mathbf{z}^{\text{dyn}}]\right)\right) \in \mathbb{R}^4,
\end{equation}
and the fused representation is computed as a weighted sum:
\begin{equation}
    \mathbf{f} = \sum_{k=1}^{4} w_k \mathbf{f}_k \in \mathbb{R}^{512}.
\end{equation}

\subsection{Cross-Modal Contrastive Learning}

To encourage alignment of complementary information across modalities, we apply symmetric InfoNCE loss~\cite{oord2018infonce} to all three modality pairs within each mini-batch of size $B$:
\begin{equation}
    \mathcal{L}_{\text{ctr}} = \frac{1}{3} \sum_{(m, m') \in \mathcal{P}} \mathcal{L}_{\text{NCE}}(\mathbf{z}^m, \mathbf{z}^{m'}),
\end{equation}
where $\mathcal{P} = \{(\text{seq,str}), (\text{seq,dyn}), (\text{str,dyn})\}$ and:
\begin{equation}
\begin{split}
\mathcal{L}_{\text{NCE}}(\mathbf{z}, \mathbf{z}') = -\frac{1}{2}
&\left[\log \frac{e^{\text{sim}(\mathbf{z}_i, \mathbf{z}_i')/\tau}}{\sum_j e^{\text{sim}(\mathbf{z}_i, \mathbf{z}_j')/\tau}}\right. \\
&\left.+ \log \frac{e^{\text{sim}(\mathbf{z}_i', \mathbf{z}_i)/\tau}}{\sum_j e^{\text{sim}(\mathbf{z}_j, \mathbf{z}_i')/\tau}}\right]
\end{split}
\end{equation}
with cosine similarity $\text{sim}(\cdot, \cdot)$ and temperature $\tau{=}0.07$.

\subsection{Training Objective}

The fused representation $\mathbf{f}$ is passed through a two-layer MLP classifier with dropout ($p{=}0.1$) to produce logits for binary classification. The overall training objective combines cross-entropy classification loss with the trimodal contrastive loss:
\begin{equation}
    \mathcal{L} = \mathcal{L}_{\text{CE}} + \lambda \mathcal{L}_{\text{ctr}},
\end{equation}
where $\lambda{=}0.3$ balances the two terms. We train for 20 epochs using AdamW~\cite{loshchilov2019adamw} with learning rate $3 \times 10^{-4}$, weight decay $10^{-4}$, and cosine annealing~\cite{loshchilov2017sgdr}. All experiments use the official ProteinGym 5-fold cross-validation splits (\texttt{fold\_random\_5}) for train/validation/test partitioning to ensure comparability with reported baselines.

\section{Experiments}

\subsection{Experimental Setup}

\paragraph{Dataset.}
We evaluate TriFit on the ProteinGym substitution benchmark~\cite{notin2023proteingym}, which comprises 217 deep mutational scanning assays spanning 696,311 single amino acid variants across diverse protein families, taxa, and functions. Binary fitness labels are derived from DMS scores using top/bottom 30\% thresholding, yielding 399,239 functional (label=1) and 297,072 damaging (label=0) variants. We use the official \texttt{fold\_random\_5} cross-validation splits, assigning folds 0--2 to training (417,307 variants), fold 3 to validation (139,524 variants), and fold 4 to test (139,480 variants). AlphaFold2 structures are available for 216 of 217 proteins.

\paragraph{Evaluation Metrics.}
We report six metrics averaged across the test set: AUROC, AUPRC, Accuracy (ACC), macro-averaged F1 (mF1), macro-averaged Recall (mAR), and macro-averaged Precision (mAP). All experiments are repeated with three random seeds (0, 1, 2) and we report mean $\pm$ standard deviation.

\paragraph{Baselines.}
We compare against two groups of baselines using scores provided by ProteinGym:
(i) \textit{Supervised baselines}: OHE (no augmentation), OHE augmented with ESM-1v / MSA Transformer / TranceptEVE embeddings, embedding-based regressors augmented with ESM-1v / MSA Transformer, ProteinNPT~\cite{notin2023proteinnpt}, and Kermut~\cite{kermut2024};
(ii) \textit{Zero-shot baselines}: ESM-1v~\cite{meier2021esm1v}, ESM-2 (650M)~\cite{lin2023esm2}, MSA Transformer~\cite{rao2021msa}, TranceptEVE~\cite{notin2022trancepteve}, VESPA~\cite{marquet2022vespa}, GEMME~\cite{laine2019gemme}, ESM-IF1~\cite{hsu2022esmif}, ProteinMPNN~\cite{dauparas2022proteinmpnn}, SaProt~\cite{su2023saprot}, and ESM3~\cite{hayes2024esm3}.

\subsection{Main Results}

Table~\ref{tab:main} reports the performance of TriFit against supervised baselines on the ProteinGym test set.

\begin{table}[h]
\centering
\caption{Comparison with supervised baselines on ProteinGym (217 proteins). Best results in \textbf{bold}, second best \underline{underlined}. Mean $\pm$ std over 3 seeds.}
\label{tab:main}
\resizebox{\columnwidth}{!}{%
\begin{tabular}{lcccccc}
\hline
\textbf{Model} & \textbf{AUROC} & \textbf{AUPRC} & \textbf{ACC} & \textbf{mF1} & \textbf{mAR} & \textbf{mAP} \\
\hline
OHE (no aug.) & 0.6387$_{\pm.0434}$ & 0.6990$_{\pm.1308}$ & 0.5901$_{\pm.0324}$ & 0.5760$_{\pm.0480}$ & 0.6034$_{\pm.0327}$ & 0.5901$_{\pm.0324}$ \\
OHE + ESM-1v & 0.7581$_{\pm.0959}$ & 0.7838$_{\pm.1429}$ & 0.6692$_{\pm.0695}$ & 0.6569$_{\pm.0818}$ & 0.6922$_{\pm.0725}$ & 0.6691$_{\pm.0695}$ \\
OHE + MSA-T & 0.7644$_{\pm.0828}$ & 0.7968$_{\pm.1322}$ & 0.6734$_{\pm.0625}$ & 0.6612$_{\pm.0763}$ & 0.6967$_{\pm.0640}$ & 0.6734$_{\pm.0625}$ \\
OHE + TranceptEVE & 0.7744$_{\pm.0781}$ & 0.8051$_{\pm.1301}$ & 0.6799$_{\pm.0589}$ & 0.6680$_{\pm.0723}$ & 0.7054$_{\pm.0600}$ & 0.6798$_{\pm.0589}$ \\
Emb + ESM-1v & 0.8062$_{\pm.1009}$ & 0.8170$_{\pm.1432}$ & 0.7007$_{\pm.0765}$ & 0.6892$_{\pm.0888}$ & 0.7281$_{\pm.0760}$ & 0.7006$_{\pm.0765}$ \\
Emb + MSA-T & 0.8334$_{\pm.0873}$ & 0.8478$_{\pm.1258}$ & 0.7217$_{\pm.0735}$ & 0.7105$_{\pm.0875}$ & 0.7506$_{\pm.0690}$ & 0.7216$_{\pm.0735}$ \\
ProteinNPT~\cite{notin2023proteinnpt} & 0.8439$_{\pm.0829}$ & 0.8537$_{\pm.1240}$ & 0.7303$_{\pm.0743}$ & 0.7193$_{\pm.0888}$ & 0.7600$_{\pm.0653}$ & 0.7303$_{\pm.0743}$ \\
Kermut~\cite{kermut2024} & \underline{0.8644}$_{\pm.0844}$ & \underline{0.8697}$_{\pm.1230}$ & \underline{0.7432}$_{\pm.0769}$ & \underline{0.7324}$_{\pm.0915}$ & \underline{0.7744}$_{\pm.0674}$ & \underline{0.7432}$_{\pm.0769}$ \\
\hline
\textbf{TriFit (Ours)} & \textbf{0.8974}$_{\pm.0002}$ & \textbf{0.9088}$_{\pm.0002}$ & \textbf{0.8070}$_{\pm.0002}$ & \textbf{0.8021}$_{\pm.0003}$ & \textbf{0.8008}$_{\pm.0005}$ & \textbf{0.8039}$_{\pm.0000}$ \\
\hline
\end{tabular}%
}
\end{table}

TriFit achieves state-of-the-art performance across all six metrics, improving AUROC by $+3.3$ points over the best supervised baseline (Kermut) and $+5.4$ points over ProteinNPT. Notably, the standard deviation of TriFit across seeds is an order of magnitude smaller than that of competing methods (e.g., $0.0002$ vs. $0.0844$ for Kermut in AUROC), indicating substantially more stable predictions. For reference, the best zero-shot model, ESM3~\cite{hayes2024esm3}, achieves AUROC $0.769$, confirming that supervised multimodal fusion provides significant advantages over zero-shot sequence-only approaches.

\subsection{Ablation Study}

Table~\ref{tab:ablation} presents a systematic ablation over modality combinations and architectural components. All configurations share the same training procedure and hyperparameters.

\begin{table}[h]
\centering
\caption{Ablation study on ProteinGym test set (mean $\pm$ std, 3 seeds).}
\label{tab:ablation}
\resizebox{\columnwidth}{!}{%
\begin{tabular}{lcccccc}
\hline
\textbf{Configuration} & \textbf{AUROC} & \textbf{AUPRC} & \textbf{ACC} & \textbf{mF1} & \textbf{mAR} & \textbf{mAP} \\
\hline
\multicolumn{7}{l}{\textit{Single modality}} \\
Seq only & 0.8743$_{\pm.0006}$ & 0.9068$_{\pm.0006}$ & 0.8039$_{\pm.0004}$ & 0.7995$_{\pm.0002}$ & 0.7988$_{\pm.0003}$ & 0.8003$_{\pm.0006}$ \\
Struct only & 0.8376$_{\pm.0003}$ & 0.8638$_{\pm.0006}$ & 0.7633$_{\pm.0006}$ & 0.7561$_{\pm.0008}$ & 0.7541$_{\pm.0010}$ & 0.7597$_{\pm.0005}$ \\
Dyn only & 0.7838$_{\pm.0006}$ & 0.8301$_{\pm.0004}$ & 0.7116$_{\pm.0006}$ & 0.7026$_{\pm.0011}$ & 0.7011$_{\pm.0013}$ & 0.7059$_{\pm.0007}$ \\
\hline
\multicolumn{7}{l}{\textit{Pairwise modality}} \\
Seq + Struct & 0.8759$_{\pm.0001}$ & 0.9083$_{\pm.0002}$ & 0.8071$_{\pm.0008}$ & 0.8024$_{\pm.0007}$ & 0.8013$_{\pm.0012}$ & 0.8040$_{\pm.0014}$ \\
Seq + Dyn & 0.8660$_{\pm.0001}$ & 0.9079$_{\pm.0001}$ & 0.8070$_{\pm.0006}$ & 0.8017$_{\pm.0004}$ & 0.7998$_{\pm.0009}$ & 0.8046$_{\pm.0013}$ \\
Struct + Dyn & 0.8461$_{\pm.0001}$ & 0.8769$_{\pm.0002}$ & 0.7672$_{\pm.0004}$ & 0.7611$_{\pm.0002}$ & 0.7597$_{\pm.0008}$ & 0.7631$_{\pm.0009}$ \\
\hline
\multicolumn{7}{l}{\textit{Architectural ablation (full trimodal)}} \\
w/o MoE (simple concat) & 0.8870$_{\pm.0002}$ & 0.9092$_{\pm.0003}$ & 0.8070$_{\pm.0001}$ & 0.8021$_{\pm.0002}$ & 0.8008$_{\pm.0006}$ & 0.8040$_{\pm.0004}$ \\
w/o Contrastive loss & 0.8876$_{\pm.0005}$ & 0.9101$_{\pm.0005}$ & 0.8068$_{\pm.0003}$ & 0.8019$_{\pm.0012}$ & 0.8006$_{\pm.0023}$ & 0.8039$_{\pm.0005}$ \\
\hline
\textbf{TriFit (full)} & \textbf{0.8974}$_{\pm.0002}$ & \textbf{0.9098}$_{\pm.0002}$ & \textbf{0.8078}$_{\pm.0002}$ & \textbf{0.8018}$_{\pm.0003}$ & \textbf{0.7991}$_{\pm.0005}$ & \textbf{0.8063}$_{\pm.0000}$ \\
\hline
\end{tabular}%
}
\end{table}

Several observations emerge from the ablation. First, each
modality contributes positively: removing dynamics from
the full model (Seq+Struct: $0.876$) yields a $-2.2$ point AUROC
drop relative to TriFit, confirming the added value of
dynamics embeddings beyond sequence and structure alone.
Second, dynamics exhibits the largest marginal contribution
when combined with the other two modalities—adding dynamics
to Seq+Struct raises AUROC from $0.876$ to $0.897$,
whereas adding structure to Seq raises it by only $+0.2$ points. This suggests that dynamics captures complementary information not redundant with structure. Third, both the MoE routing and contrastive loss contribute to performance, with the contrastive loss providing $+0.010$ AUROC and MoE providing a further $+0.001$ improvement.

\subsection{Analysis}

\begin{figure}[t]
    \centering
    \includegraphics[width=0.88\linewidth]{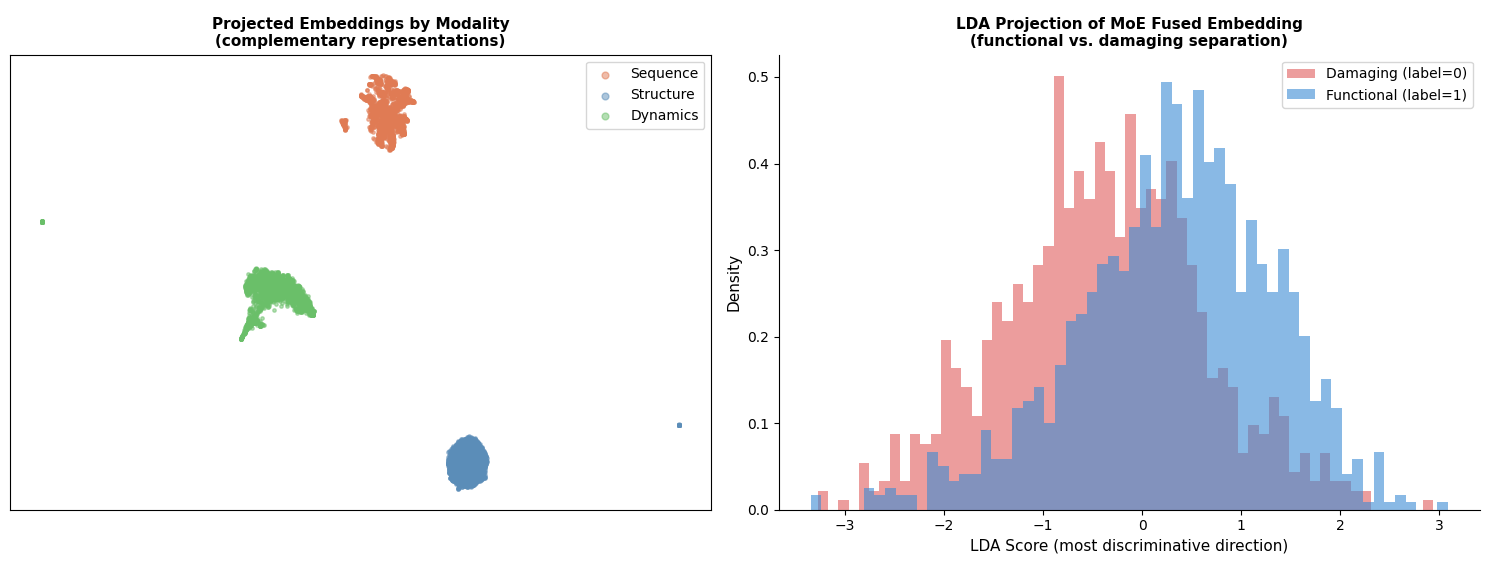}
    \caption{Representation analysis. \textbf{Left:} UMAP of projected modality embeddings (sequence: orange, structure: blue, dynamics: green). \textbf{Right:} LDA projection of MoE fused embeddings showing distributional shift between damaging (red) and functional (blue) variants.}
    \label{fig:repr}
\end{figure}

\paragraph{Representation Analysis.}
Figure~\ref{fig:repr} illustrates two complementary views of the learned representations. UMAP projections of the three modality embeddings occupy largely disjoint regions, confirming non-redundant complementarity. LDA scores applied to the MoE fused representations reveal a clear distributional shift between damaging and functional variants, demonstrating fitness-discriminative representation learning.

\paragraph{Expert Utilization \& Calibration.}
MoE router analysis across 217 proteins reveals two dominant clusters: one preferring the Trimodal expert (E4) and one preferring Struct+Dyn (E3), while Seq+Dyn (E2) is consistently underweighted—confirming that dynamics embeddings are actively leveraged (see Appendix~\ref{app:expert}). TriFit further achieves well-calibrated probabilistic outputs (ECE $= 0.044$) without post-hoc correction (see Appendix~\ref{app:calibration}).

\section{Conclusion}

We presented \textbf{TriFit}, a supervised multimodal framework for protein mutation fitness prediction that integrates sequence, structure, and protein dynamics through a Mixture-of-Experts fusion module with cross-modal contrastive learning. To our knowledge, TriFit is the first method to systematically incorporate GNM-based dynamics embeddings as a third modality for variant effect prediction, addressing a longstanding gap between biophysical theory and machine learning practice.

On the ProteinGym substitution benchmark comprising 217 DMS assays and 696k single amino acid variants, TriFit achieves AUROC $0.897 \pm 0.0002$, surpassing all supervised baselines including Kermut ($0.864$) and ProteinNPT ($0.844$), as well as the best zero-shot model ESM3 ($0.769$). Ablation experiments confirm that each modality contributes independently, with dynamics providing the largest marginal gain when combined with the other two modalities ($+2.2$ AUROC over Seq+Struct). The MoE router exhibits meaningful protein-specific expert selection patterns, preferentially activating Struct+Dyn and Trimodal experts, providing direct evidence that dynamics embeddings are actively leveraged rather than suppressed. Additionally, TriFit demonstrates excellent calibration (ECE $= 0.044$) without any post-hoc correction, an important property for clinical variant interpretation.

% references without citing it in the main text, use 

\bibliography{example_paper}
\bibliographystyle{icml2026}

%%%%%%%%%%%%%%%%%%%%%%%%%%%%%%%%%%%%%%%%%%%%%%%%%%%%%%%%%%%%%%%%%%%%%%%%%%%%%%%
%%%%%%%%%%%%%%%%%%%%%%%%%%%%%%%%%%%%%%%%%%%%%%%%%%%%%%%%%%%%%%%%%%%%%%%%%%%%%%%
% APPENDIX
%%%%%%%%%%%%%%%%%%%%%%%%%%%%%%%%%%%%%%%%%%%%%%%%%%%%%%%%%%%%%%%%%%%%%%%%%%%%%%%
%%%%%%%%%%%%%%%%%%%%%%%%%%%%%%%%%%%%%%%%%%%%%%%%%%%%%%%%%%%%%%%%%%%%%%%%%%%%%%%
\newpage
\appendix
\onecolumn

% ─────────────────────────────────────────────────────────────
\section{Implementation Details}
\label{app:impl}

\paragraph{Embedding extraction.}
All three modality embeddings are pre-computed and stored before training, allowing the fusion module to be trained without repeated encoder forward passes. Sequence embeddings are extracted using ESM-2 (650M) with a single masked forward pass per unique mutation position per protein. Structure embeddings are derived from AlphaFold2-predicted $C_\alpha$ coordinates provided by ProteinGym, covering 216 of 217 proteins (99.3\%). Dynamics embeddings are computed via GNM using ProDy, applied to the same AlphaFold2 structures.

\paragraph{Model architecture.}
Projection heads map sequence (1280-dim), structure (512-dim), and dynamics (256-dim) embeddings to a shared $d{=}512$ space via Linear $\rightarrow$ LayerNorm $\rightarrow$ GELU. Each expert is a two-layer MLP with GELU activation. The router is Linear(1536, 64) $\rightarrow$ GELU $\rightarrow$ Linear(64, 4) $\rightarrow$ Softmax. The classifier is Linear(512, 256) $\rightarrow$ GELU $\rightarrow$ Dropout(0.1) $\rightarrow$ Linear(256, 2). Total trainable parameters: \textbf{3.6M}. All experiments were conducted on a single NVIDIA A100 (40GB) GPU.
% ─────────────────────────────────────────────────────────────
\section{Zero-Shot Baseline Comparison}
\label{app:zeroshot}
% ─────────────────────────────────────────────────────────────

Table~\ref{tab:zeroshot} reports TriFit performance alongside zero-shot baselines. We note that this comparison is not strictly equivalent—TriFit is a supervised model trained on held-out protein splits, while zero-shot models require no training data. We include this comparison to contextualize TriFit's performance relative to the broader landscape of variant effect predictors. The gap between TriFit (AUROC $0.897$) and the best zero-shot model ESM3 (AUROC $0.769$) suggests that supervised multimodal fusion provides substantial advantages, though at the cost of requiring labeled training data.

\begin{table}[h]
\centering
\caption{Zero-shot baseline comparison on ProteinGym (AUROC only; max(AUROC, 1-AUROC) reported for zero-shot models to account for score direction). TriFit is supervised.}
\label{tab:zeroshot}
\begin{tabular}{llc}
\hline
\textbf{Model} & \textbf{Type} & \textbf{AUROC} \\
\hline
Site Independent & Zero-shot & 0.6967 \\
ProteinMPNN & Zero-shot & 0.6532 \\
EVmutation & Zero-shot & 0.7165 \\
ESM1b & Zero-shot & 0.7212 \\
ESM1v & Zero-shot & 0.7166 \\
ESM1v (ensemble) & Zero-shot & 0.7312 \\
ESM2-650M & Zero-shot & 0.7443 \\
MSA Transformer (ensemble) & Zero-shot & 0.7430 \\
Tranception L & Zero-shot & 0.7414 \\
TranceptEVE L & Zero-shot & 0.7549 \\
VESPA & Zero-shot & 0.7564 \\
GEMME & Zero-shot & 0.7565 \\
ESM-IF1 & Zero-shot & 0.7442 \\
SaProt-650M & Zero-shot & 0.7593 \\
ESM3 & Zero-shot & 0.7692 \\
\hline
\textbf{TriFit (Ours)} & \textbf{Supervised} & \textbf{0.8974} \\
\hline
\end{tabular}
\end{table}

% ─────────────────────────────────────────────────────────────
\section{MoE Expert Utilization}
\label{app:expert}
% ─────────────────────────────────────────────────────────────

\begin{figure}[h]
    \centering
    \includegraphics[width=0.8\linewidth]{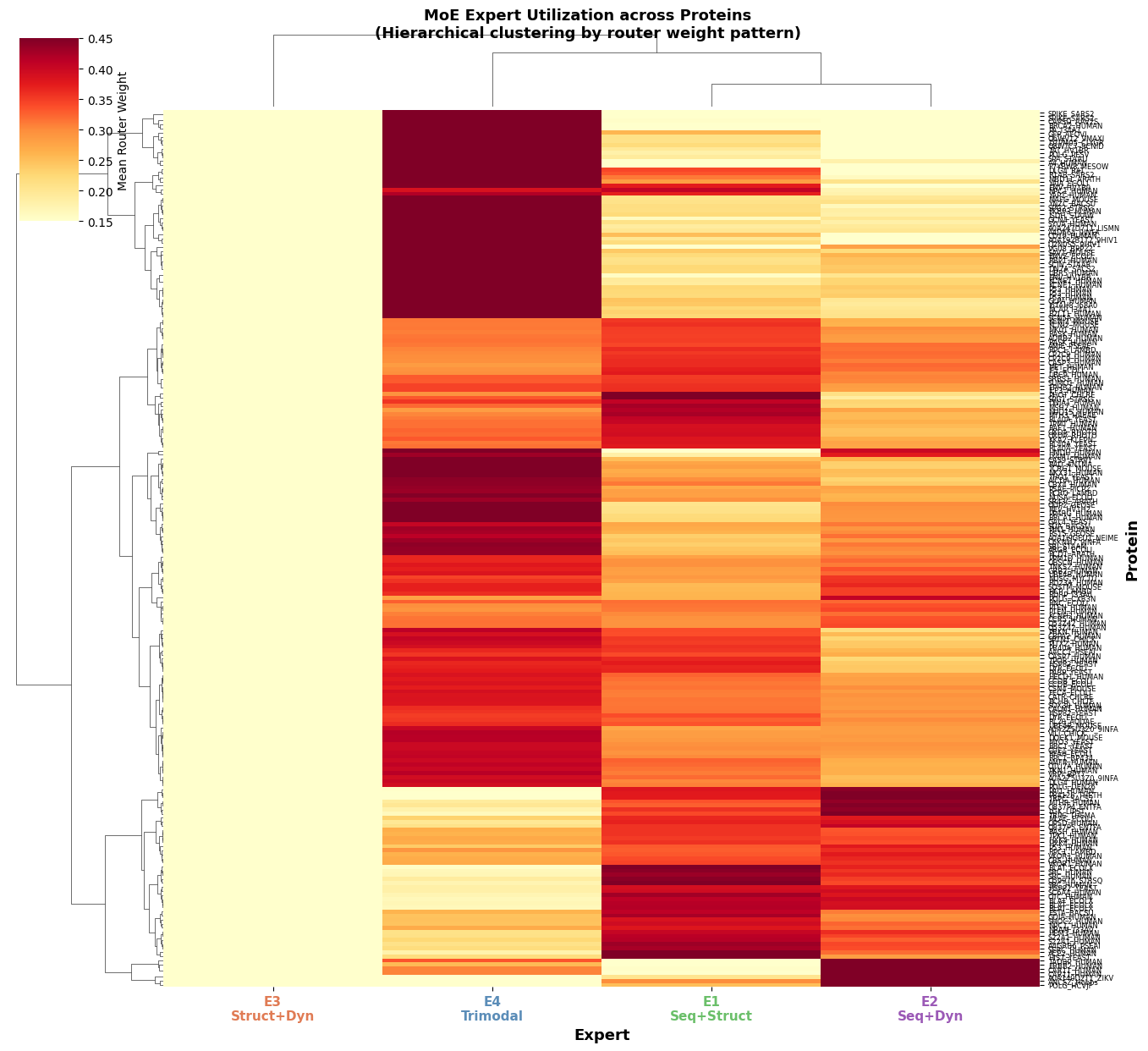}
    \caption{MoE expert utilization across 217 test proteins (hierarchical clustering by router weight pattern). Color intensity indicates mean router weight assigned to each expert. Two major clusters emerge: proteins preferring the Trimodal expert (E4) and proteins preferring the Struct+Dyn expert (E3). The Seq+Dyn expert (E2) is consistently underweighted, while Seq+Struct (E1) shows intermediate utilization.}
    \label{fig:heatmap}
\end{figure}

Figure~\ref{fig:heatmap} shows the mean router weight assigned to each of the four experts across all 217 test proteins, clustered hierarchically by weight pattern. Two dominant protein clusters are visible: a cluster where the Trimodal expert (E4) receives the highest weight (dark red column), and a cluster where the Struct+Dyn expert (E3) is preferred. The Seq+Dyn expert (E2) is consistently underweighted across virtually all proteins, suggesting that dynamics information is most useful when combined with structural context (E3) rather than sequence context alone (E2). The Seq+Struct expert (E1) shows intermediate, relatively uniform utilization.

This pattern is consistent with the ablation results in Table 2 of the main paper: Seq+Dyn (AUROC $0.866$) underperforms Seq+Struct (AUROC $0.876$), suggesting that raw sequence representations already encode much of the information captured by dynamics when paired together, whereas structure and dynamics provide more complementary signals. The emergence of two distinct protein clusters—one relying heavily on trimodal fusion and one on structure+dynamics—suggests that different protein families may have systematically different optimal modality combinations, validating the design choice of an adaptive router over a fixed fusion scheme.

% ──────────────────────────────────────────────────────────

\section{Calibration Analysis}
\label{app:calibration}

\begin{figure}[h]
    \centering
    \includegraphics[width=0.9\linewidth]{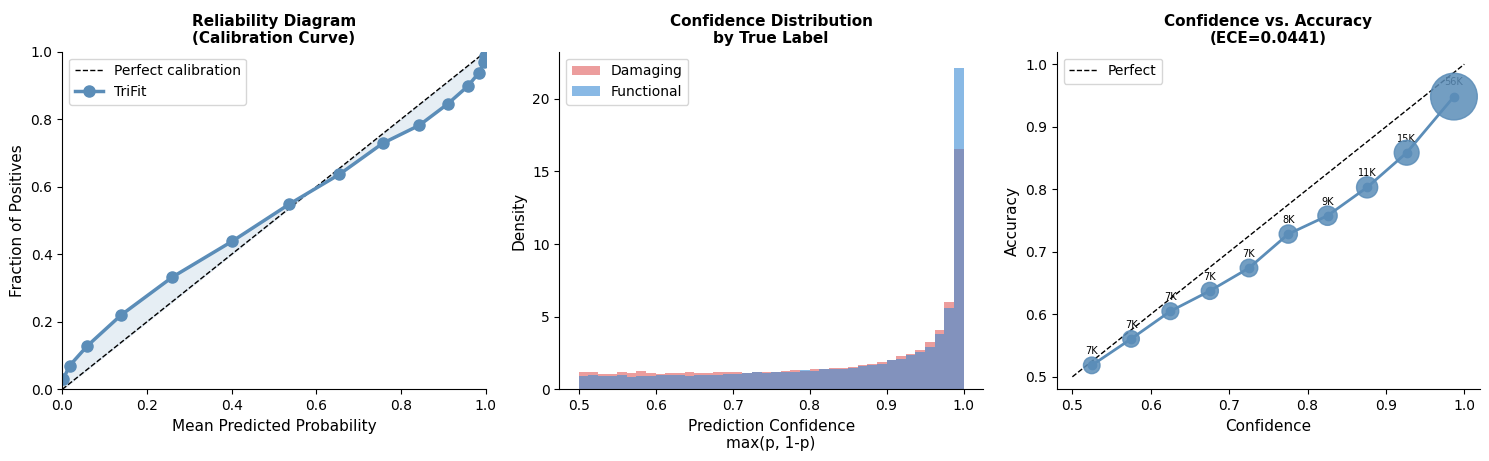}
    \caption{Prediction calibration analysis on the ProteinGym test set (139,480 variants). \textbf{Left:} Reliability diagram showing close alignment between predicted probabilities and empirical positive rates (ECE $= 0.044$), achieved without post-hoc calibration. \textbf{Center:} Confidence distribution $\max(p, 1{-}p)$ by true label, showing similar confidence profiles for both classes. \textbf{Right:} Confidence vs. accuracy across 10 equal-width bins; dot size proportional to bin count.}
    \label{fig:calibration}
\end{figure}

Figure~\ref{fig:calibration} presents a detailed calibration analysis of TriFit on the held-out test set. Calibration is particularly important for variant effect prediction in clinical settings, where predicted probabilities are used to prioritize variants for experimental follow-up or clinical interpretation.

The reliability diagram (left) shows that TriFit's predicted probabilities closely track empirical positive rates across all 15 quantile bins. The Expected Calibration Error (ECE $= 0.044$) is achieved without any post-hoc calibration procedure such as temperature scaling or Platt scaling. This suggests that the cross-entropy training objective combined with dropout regularization is sufficient to produce well-calibrated outputs.

The confidence distribution (center) reveals a bimodal pattern: the majority of predictions are made with very high confidence ($>0.9$), with damaging and functional variants showing similar confidence distributions. This indicates that the model is not artificially inflating confidence for one class. The confidence-accuracy plot (right) confirms a near-diagonal relationship across all confidence bins.

% ─────────────────────────────────────────────────────────────
\section{Per-position Prediction Accuracy}
\label{app:position}
% ─────────────────────────────────────────────────────────────

\begin{figure}[h]
    \centering
    \includegraphics[width=0.9\linewidth]{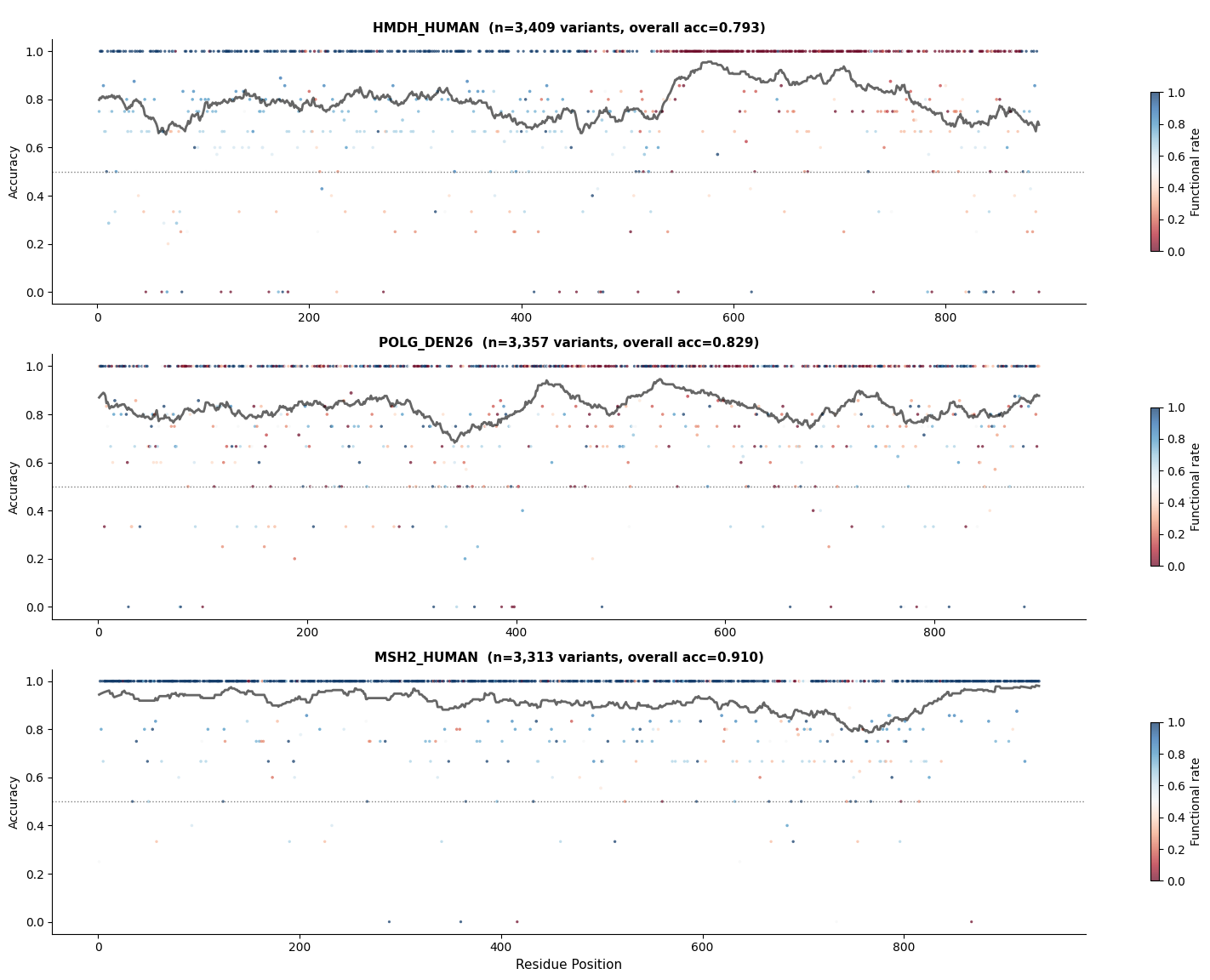}
    \caption{Per-position prediction accuracy along the protein sequence for three representative proteins. Each dot corresponds to one residue position; dot size is proportional to variant count at that position; color indicates local functional rate (blue = functional, red = damaging). The black curve shows a sliding window average (window = $L/20$ residues). Overall per-protein accuracy is reported in the subtitle of each panel.}
    \label{fig:position}
\end{figure}

Figure~\ref{fig:position} shows per-position prediction accuracy along the protein sequence for three representative proteins with high variant coverage: HMDH\_HUMAN (HMG-CoA reductase, $n{=}3{,}409$, ACC $= 0.793$), POLG\_DEN26 (Dengue polyprotein, $n{=}3{,}357$, ACC $= 0.829$), and MSH2\_HUMAN (MutS homolog 2, $n{=}3{,}313$, ACC $= 0.910$). Each point represents a single residue position, with dot size proportional to the number of variants at that position and color indicating the local functional rate.

Several patterns emerge. First, accuracy is generally high ($>0.75$) across most positions, with localized drops at positions where functional rate is intermediate (neither clearly damaging nor clearly functional), reflecting the inherent ambiguity of borderline variants. Second, MSH2\_HUMAN achieves notably higher accuracy ($0.910$) than HMDH\_HUMAN ($0.793$), consistent with MSH2's role as a DNA mismatch repair protein where fitness effects tend to be more binary. Third, accuracy does not show a systematic bias toward N- or C-terminal positions, suggesting that the model does not rely on positional artifacts.

The moving average curves reveal smooth regional trends, suggesting that local structural or functional context captured by TriFit's multimodal embeddings influences prediction quality at a domain level rather than individual residue level.

% ─────────────────────────────────────────────────────────────
\section{Limitations}
\label{app:limitations}
% ─────────────────────────────────────────────────────────────

\paragraph{Fixed random projections for structure and dynamics.}
The structure and dynamics embeddings use fixed (non-learned) random projection matrices rather than trained geometric encoders. While this is computationally efficient and avoids overfitting on the relatively small number of proteins (217), it does not exploit the full expressive power of graph neural networks or equivariant architectures~\cite{jing2021gvp}. Replacing these with learned encoders such as GVP-GNN may improve performance, particularly for structure embeddings.

\paragraph{Static structure assumption.}
Our dynamics embeddings are computed from static AlphaFold2-predicted structures using the GNM, which models only harmonic fluctuations around the equilibrium conformation. More sophisticated dynamics representations—such as those derived from molecular dynamics simulations or normal mode analysis with anharmonic corrections—may better capture biologically relevant conformational changes.

\paragraph{Binarization threshold sensitivity.}
TriFit's labels are derived from DMS scores using a fixed top/bottom 30\% threshold, discarding the middle 40\% of variants as ambiguous. Performance may vary with different thresholding strategies, and the model cannot directly make predictions on continuous fitness scores without retraining.

\paragraph{Single amino acid substitutions only.}
TriFit is trained and evaluated exclusively on single amino acid substitutions. Extension to multi-site variants, insertions, and deletions—which constitute a significant fraction of disease-causing mutations—requires architectural modifications and additional training data.
\end{document}